\pdfoutput=1

\documentclass[11pt]{article}
\usepackage{acl}

\usepackage{amsmath, amssymb, algorithm, algorithmic}
\usepackage{changepage}
\usepackage{xspace, mfirstuc, tabulary}
\definecolor{indomain}{RGB}{144,238,144} 
\definecolor{inndomain}{RGB}{253,127,57} 
\definecolor{outdomain}{RGB}{173,216,230} 
\definecolor{lightgray}{gray}{0.9}

\usepackage{comment}
\usepackage{times}
\usepackage{latexsym}
\usepackage{graphicx}
\usepackage{float}
\usepackage[T1]{fontenc}
\usepackage[utf8]{inputenc}
\usepackage{microtype}
\usepackage{booktabs}
\usepackage{multicol}
\usepackage{multirow}
\usepackage{tcolorbox}
\usepackage{enumitem}
\usepackage{amsfonts}
\usepackage{array}
\usepackage{tabularx}
\usepackage{verbatim}
\usepackage{colortbl}
\usepackage{arydshln}
\usepackage{subcaption}
\usepackage{longtable}
\usepackage{placeins} 
\usepackage{xcolor}
\usepackage{listings}
\usepackage{soul}
\usepackage{mathrsfs}
\usepackage{supertabular}
\usepackage{dsfont}
\usepackage{geometry}
\usepackage{url}
\definecolor{pastelgreen}{HTML}{97D077}
\definecolor{customisedblue}{HTML}{6C8EBF}
\definecolor{lightblue}{HTML}{DAE8FC}

\definecolor{citecolor}{RGB}{34,139,34}

\newcommand{\bcol}[1]{\fontsize{8.5}{42}\selectfont\textbf{(#1)}}

\newcommand{\gcol}[1]{{\bf \fontsize{8.5}{42}\selectfont \color{citecolor!80}~(#1)}}

\newcommand{\QuaSAR}{\emph{QuaSAR}\xspace}

\title{Improving Chain-of-Thought Reasoning via Quasi-Symbolic Abstractions}

\author{\textbf{Leonardo Ranaldi$^{\dagger,\oplus}$ \qquad  Marco Valentino$^{\dagger,\ast}$ \qquad  Andr\'e Freitas$^{\dagger,\circ,\bullet}$} \\
	$^{\dagger}$Idiap Research Institute, Switzerland \\
    $^{\oplus}$School of Informatics, University of Edinburgh, UK\\
	$^{\ast}$School of Computer Science, University of Sheffield, UK\\
 $^{\circ}$Department of Computer Science, University of Manchester, UK \\
 $^{\bullet}$National Biomarker Centre (NBC), CRUK Manchester Institute, UK\\
		{
  {\tt [name].[surname]@idiap.ch}
  } }

\begin{document}
\maketitle
\begin{abstract}
Chain-of-Thought (CoT) represents a common strategy for reasoning in Large Language Models (LLMs) by decomposing complex tasks into intermediate inference steps. However, explanations generated via CoT are susceptible to \emph{content biases} that negatively affect their \emph{robustness} and \emph{faithfulness}.
To mitigate existing limitations, recent work has proposed the use of logical formalisms coupled with external symbolic solvers. However, fully symbolically formalised approaches introduce the bottleneck of requiring a complete translation from natural language to formal languages, a process that affects efficiency and flexibility.
To achieve a trade-off, this paper investigates methods to disentangle content from logical reasoning without a complete formalisation. In particular, we present \QuaSAR (for \underline{Qua}si-\underline{S}ymbolic \underline{A}bstract \underline{R}easoning), a variation of CoT that guides LLMs to operate at a higher level of abstraction via \emph{quasi-symbolic explanations}. Our framework leverages the capability of LLMs to formalise only relevant variables and predicates, enabling the coexistence of symbolic elements with natural language. We show the impact of \QuaSAR for in-context learning and for constructing demonstrations to improve the reasoning capabilities of smaller models. Our experiments show that quasi-symbolic abstractions can improve CoT-based methods by up to 8\% accuracy, enhancing robustness and consistency on challenging adversarial variations on both natural language (i.e. MMLU-Redux) and symbolic reasoning tasks (i.e., GSM-Symbolic).
\end{abstract}

\section{Introduction}

\textit{Multi-step reasoning methods}, best exemplified by Chain-of-Thought \cite{wei2022chain,wang2022self}, have been proposed to improve the performance of Large Language Models (LLMs) on downstream tasks by breaking down complex problems into intermediate reasoning steps. The success of these methods is due to the LLMs’ properties of performing tasks by following in-context structured requirements \cite{zhou2023leasttomostpromptingenablescomplex,dong2024surveyincontextlearning,ranaldi-etal-2024-tree,ranaldi-etal-2024-empowering-multi}.

Despite CoT being the current workhorse for LLM reasoning, complex reasoning still remains a significant challenge for LLMs \cite{10.1162/tacl_a_00594,luo2024improve}, with recent work showing that explanations generated via CoT are susceptible to \emph{content biases} that negatively affect their \emph{robustness} and \emph{faithfulness} \cite{lyu2023faithful,turpin2024language,yee2024dissociation}.

To mitigate these limitations and improve reasoning capabilities, recent works have proposed using logical formalisms \cite{lyu2023faithful,jiang2024leanreasoner,arakelyan2024flarefaithfullogicaidedreasoning}. However, fully symbolic approaches possess the bottleneck of requiring a complete translation from natural language to formal languages, a process that negatively impacts efficiency and flexibility \cite{dinh2023largelanguagemodelscode,quan2024verificationrefinementnaturallanguage,quan-etal-2024-enhancing,dalal-etal-2024-inference}.

\begin{figure*}[t]
\centering
    \includegraphics[width=\textwidth]{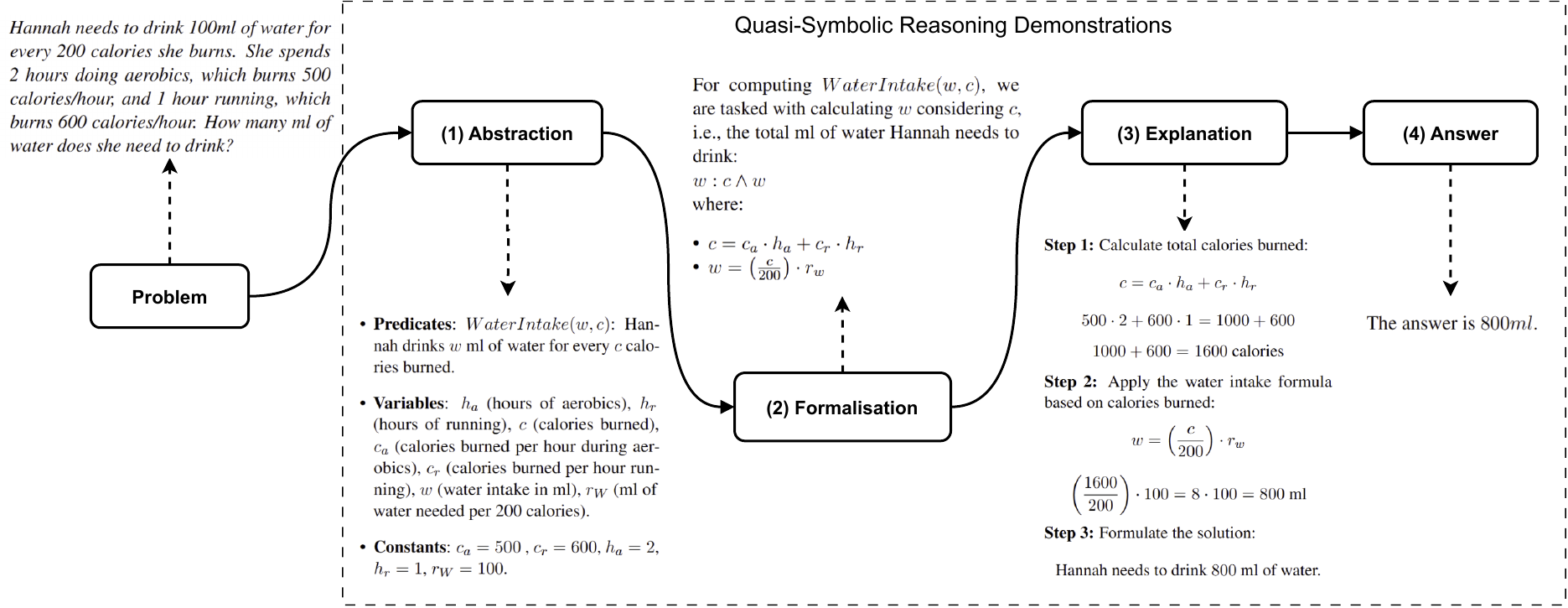}
    \caption{\QuaSAR elicits quasi-symbolic abstractions in LLMs via the following steps: \textit{(i)} \textit{Abstraction}, where the problem is analysed and abstracted in terms of relevant symbolic predicates, variables, and constants; \textit{(ii)} \textit{Formalisation}, where the original problem is reformulated using a mixture of symbolic expressions and natural language; \textit{(iii)} \textit{Explanation}, where the necessary steps to compute the solution are formulated via quasi-symbolic reasoning chains; and \textit{(iv)} \textit{Answering}, where a final solution is generated. We use \QuaSAR as an in-context learning strategy and for constructing reasoning demonstrations for smaller LLMs. }
    \label{fig:pipeline}
\end{figure*}

This paper investigates methods to achieve a better trade-off between flexibility and robustness by disentangling content from logical reasoning without the need for a complete formalisation. In particular, we present \emph{QuaSAR} (for \underline{Qua}si-\underline{S}ymbolic \underline{A}bstract \underline{R}easoning), a variation to CoT that guides LLMs to operate at a higher level of abstraction via \emph{quasi-symbolic explanations}. Our framework leverages the capability of LLMs to formalise relevant variables and predicates, enabling the coexistence of symbolic elements with natural language.

Specifically, the aim of \QuaSAR is to enable LLMs in tackling complex multi-step reasoning problems via the following steps: \textit{(i)} \textit{Abstraction}, where the problem is analysed and abstracted in terms of relevant symbolic predicates, variables, and constants; \textit{(ii)} \textit{Formalisation}, where the original problem is reformulated using a combination of a minimal symbolic form and natural language; \textit{(iii)} \textit{Explanation}, where the necessary steps to compute the solution are formulated via quasi-symbolic reasoning chains; and \textit{(iv)} \textit{Answering}, where a final solution is generated.

Building on recent work \cite{lyu2023faithful,jiang2024leanreasoner}, 
\QuaSAR guides LLMs via structured instructions, going beyond the problems associated with using external solvers \cite{quan2024verificationrefinementnaturallanguage}. At the same time, in contrast to work using formal languages to guide CoT reasoning \cite{leang2024comatchainmathematicallyannotated,arakelyan2024flarefaithfullogicaidedreasoning}, \QuaSAR operates via a single prompting step, reducing costs, thereby delivering robust reasoning trajectories across different types of reasoning tasks.

We demonstrate the operability of \QuaSAR in two different configurations -- as an in-context approach to provide explicit instructions for larger and more capable LLMs and as a strategy for constructing synthetic demonstrations to improve the performance and align the reasoning capabilities of smaller LLMs. Hence, we perform an extensive empirical evaluation using different LLMs (i.e., GPT-4o, Llama3, and Qwen-2) on complex mathematical problems, reasoning, and natural language understanding tasks. \QuaSAR demonstrates significant improvements by achieving an overall exact match boost on all proposed tasks.

In particular, our experiments led to the following findings and conclusions: 

\begin{enumerate}
    \item Formalising and structuring the LLMs' reasoning through quasi-symbolic trajectories enhances accuracy and verifiability, leading to an average increase in accuracy of 8\% over CoT and 6.8\% and 8.2\% over CoMAT \cite{leang2024comatchainmathematicallyannotated} and Faithful CoT \cite{lyu2023faithful} respectively when applied on GPT-4o.
    \item We found that our symbolic-inspired approach is significantly more efficient than related methods and can be employed on different tasks (i.e., mathematical and natural language reasoning tasks) without significant changes. Indeed, \QuaSAR achieves state-of-the-art performance across diverse tasks of varying complexity and languages operating through the same framework.
    \item By conducting an in-depth ablation study, we demonstrate the generalisability of \QuaSAR and its effectiveness on different scales of LLMs. Our experiments show that \QuaSAR provides more robust reasoning trajectories on tasks that are typically challenging  for smaller-scale models, enhancing robustness and consistency on challenging adversarial variations on both natural language (i.e. MMLU-Redux) and symbolic reasoning tasks (i.e., GSM-Symbolic)
\end{enumerate}

To the best of our knowledge, \QuaSAR is the first method to apply quasi-symbolic demonstrations for a broad spectrum of reasoning tasks, demonstrating the impact of enabling the co-existence of symbolic abstractions and natural language explanations for improving the efficiency and robustness of LLMs.

\section{\QuaSAR: Quasi-Symbolic Abstract Reasoning}
\label{sec:methods}

Integrating symbolic elements into natural language explanations is crucial for reasoning in disciplines such as mathematics and science, where symbolic abstractions facilitate the identification and generalisation of the logical connections between premises (i.e., explanans) and conclusions (i.e., explanandum) \cite{Wang1954,Bronkhorst2019, PENNINGTON1993123,valentino2024nature,MILLER20191}.

For example, within the unificationist account of explanation, \citet{kitcher1981explanatory} posits that explanations function by subsuming an apparently disconnected set of observations under the same underlying regularity, thereby forming recurring \emph{argument patterns}. These patterns emerge when explanations are generalised through the replacement of concrete entities and predicates with abstract symbols. This process of \emph{quasi-symbolic abstraction} enables explanatory arguments to be detached from specific world knowledge, thereby allowing their applicability across different problems (e.g., the same argument pattern created by the theory of gravity can be used to explain why specific objects fall and why celestial objects attract each other) \cite{valentino-etal-2021-unification,Valentino_Thayaparan_Ferreira_Freitas_2022,valentino-etal-2022-case,zheng2024take}.

In this paper, we aim to explicitly leverage argument patterns with LLMs, hypothesising that quasi-symbolic abstractions can help disentangle concrete world knowledge from symbolic reasoning within a natural language explanatory framework and mitigate some of the challenges related to content effect. An example of this process is illustrated in Figure \ref{fig:pipeline}.

Formally, conventional in-context reasoning methods are structured as a triplet $(\mathcal{Q}, \mathcal{R}, \mathcal{A})$, where $\mathcal{Q}$ represents the question, $\mathcal{R}$ consists of in-context multi-step reasoning explanations (expressed in natural language or a related form), and $\mathcal{A}$ denotes the final answer.
We extend this formalism by instructing the LLM to operate via explicit symbolic transformations as a core component of the reasoning process. Our framework, \QuaSAR, structures the solution process as a quadruple  $(\mathcal{Q}, \mathcal{S}, \mathcal{R}, \mathcal{A})$, where $\mathcal{S} = (s_1, s_2, s_3, s_4)$ represents a chain of instructions that guide the models to formalise relevant parts of the reasoning process. Each step $s_i$ corresponds to a structured transformation aimed at decomposing the problem into a sequence of symbolically-elicited operations. This structured decomposition enhances transparency and facilitates systematic verification of each step.

\subsection{\QuaSAR's Reasoning Process}
\label{sec:SiC_proces}

A complex problem solution could be described by a sequence of inference steps determined by identifying and isolating the problem predicates and structuring a formalisation that facilitates reasoning to reach the final solution.
Accordingly, \QuaSAR operates using four steps that aim to improve the accuracy of the reasoning trajectory in LLMs: \textit{(i)} \textit{Abstraction}, where the problem is analysed and abstracted in terms of relevant symbolic predicates, variables, and constants; \textit{(ii)} \textit{Formalisation}, where the original problem is reformulated using a mixture of symbols and natural language; \textit{(iii)} \textit{Explanation} (\S\ref{sec:Explaination}), where the transformations are solved using quasi-symbolic representations that explicitly explain the solution; and  \textit{(iv)} \textit{Answering} (\S\ref{sec:Answering}), where a final solution is generated to address the problem. Appendix \ref{app:SiC_prompt} reports \QuaSAR prompt.

\subsubsection{Abstraction}
\label{sec:Exemplification}
Abstracting the problem through the identification of relevant information is the first step in solving complex tasks and is a fundamental stage in structuring a robust formalisation \cite{Bronkhorst2019}. Therefore, as a first step, \QuaSAR instructs the LLM to exemplify predicates, variables, and constants, whether of numerical or verbal types.

\subsubsection{Formalisation}
\label{sec:Formalisation}
The crucial step of \QuaSAR is the formalisation of the problem, which aids accurate reasoning by \textit{translating} natural language into a semi-structured symbolic form. Hence, we instruct the LLM to deliver a quasi-formal representation of the problem, which is originally in natural language, using a structural-logical translation that explicitly represents the facts of the problem. This step is the basis for constructing an accurate reasoning trajectory because translating concrete terms in natural language into symbols aims to minimise ambiguities and content effects without compromising the components that may be significant for solving the problem.

\subsubsection{Explanation}
\label{sec:Explaination}
A significant component of CoT reasoning methods is breaking down the problem into a sequence of steps to arrive at the final solution.
The explanation phase is based on step-by-step reasoning \cite{kojima2022large} explicitly prompting the model or delivering in-context demonstrations, generally natural language rationales.
Then, the LLM is expected to solve the problem by providing a logical explanation that motivates the steps to the solution. In \QuaSAR, the reasoning trajectory is based on the symbolic structure. In this way \QuaSAR aims to elicit logical connections between each step, reducing the risk of errors caused by contextual knowledge or implicit symbolic-logical relations. The solution is then generated based on this quasi-symbolic reasoning process, which, although similar to the breakdown of reasoning methods, has a semi-structured formalisation standing behind it.

\subsubsection{Answering}
\label{sec:Answering}

\QuaSAR brings the reasoning trajectory to a final stage in which the LLM is instructed through a specific pattern --i.e.,  ``\textit{The answer is: [number]}''. Although not fundamental, this stage is significant as it ensures that the reasoning constructed in the previous stages has a conclusion. Furthermore, this step facilitates the evaluation as it triggers the LLM to deliver a response that follows the pattern of the evaluation task.

\subsection{\QuaSAR Application}
\label{sec:SiC_Application}

\QuaSAR leverages a set of structured instructions to deliver step-wise explanations. Thus, the operability of \QuaSAR is two-fold, as it can be used as both an in-context learning strategy and as a synthetic annotation method to support supervised learning (both described below). 

\subsubsection{\QuaSAR for In-Context Learning}
\label{sec:in-context}

Using the step described in \S \ref{sec:SiC_proces}, we adopt \QuaSAR to instruct three LLMs (i.e., GPT-4o, Llama-3-70B, and Qwen2-72B). Specifically, we instruct the models to exemplify and abstract the most important information from the given problem, formalising and translating natural language in a semi-structured logical form, explaining the solution in a step-wise manner, and finally generating the conclusive short-form answer in a strict format to have a more detailed and strict downstream evaluation. 
However, although the sequence of instructions is well-structured and defined, the ability to perform sequential and complex reasoning tasks is limited to larger LLMs (such as GPT-4-o, as discussed in the experiments). Hence, we transfer these capabilities to smaller models operating via \QuaSAR for building synthetic reasoning demonstrations as training sets.

\subsubsection{\QuaSAR for Reasoning Demonstrations}
\label{sec:annotation_strategy}
We instruct smaller models via demonstrations produced by high-performing LLMs capable of following structured instructions. 
To filter for the quality of generated demonstrations, we follow the method proposed by \citet{ranaldi-etal-2025-eliciting}, which computes the citation precision for the considered documents as a proxy for the quality of the demonstrations. However, since \QuaSAR employs a different annotation mechanism, our heuristics firstly filter out the final correct answers through a strict, exact match; then, behind the filtering (cutting off about 50\% of the demonstrations), it verifies that each retrieved document along the reference evidence has been considered (a detailed description of the annotation phase is in Appendix \ref{app:annotation}).

\subsection{Training}
\label{sec:tuning}
We train a Language Model $\theta$ using the annotations\footnote{we select annotations as described in \S \ref{sec:annotation_strategy}} generated via \QuaSAR. The annotations are augmented with reasoning demonstrations $\alpha$ using the standard language modelling objective, maximising likelihood:
\begin{equation}\label{eq:likehood}
\max _{\theta} \mathbb{E}_{(q, \alpha, y) \sim \mathcal{D}} \log p_{\theta}(Y \mid \alpha, Q) p_{\theta}({\alpha \mid Q})
\end{equation}

\noindent where $\alpha = \alpha_1 \cdot \alpha_2 \cdot \alpha_3 \cdot \alpha_4$ is the combination of the step-wise reasoning trajectory delivered by the model, ''$\cdot$'' is the concatenation operator, and $\alpha_1$, $\alpha_2$, are the respective annotations generated by the above processes. Finally, $Q$ is the provided question, and $Y$ is the answer, including the intermediate steps and the final answer. $\mathcal{D}$ is the training corpus constructed using training demonstrations.

\begin{table*}[h]
\centering
\small
\begin{tabular}{lccccccc}
\toprule
\multirow{2}{*}{\textbf{Model}} & \multicolumn{5}{c}{\textbf{Symbolic}} & \multicolumn{2}{c}{\textbf{Natural Language}} \\
\cmidrule(lr){2-6} \cmidrule(lr){7-8}
& \textbf{AQuA} & \textbf{GSM8K} & \textbf{SVAMP} & \textbf{MMLU-Redux} & \textbf{OlyBench} & \textbf{GPQA} & \textbf{DROP} \\
\midrule
GPT-4o & 72.8 & 94.0 & 90.4 & 79.7 & 9.9 & 46.5 & 83.4 \\
+ CoT & 84.3 & 94.5 & 90.3 & 88.1 & 41.8 & 50.2 & 84.2 \\
+ CoMAT \cite{leang2024comatchainmathematicallyannotated} & 83.5 & 93.7 & - & 88.3 & 40.4 & - & - \\
+ FCoT \cite{lyu2023faithful} & 73.6 & 95.0 & 95.3 & 76.8 & - & - & - \\
+ \QuaSAR & \textbf{87.4} & \textbf{96.5} & \textbf{97.0} & \textbf{90.2} & \textbf{44.6} & \textbf{55.4} & \textbf{88.9} \\
\midrule
Llama-3-70B & 70.9 & 84.9 & 79.8 & 70.8 & 14.6 & 41.3 & 81.4 \\
+ CoT & 74.0 & 86.1 & 84.6 & 82.0 & 22.8 & 41.9 & 80.2 \\
+ \QuaSAR & \textbf{79.1} & \textbf{88.2} & \textbf{84.9} & \textbf{85.7} & \textbf{38.2} & \textbf{49.2} & \textbf{88.0} \\ 
\midrule
Qwen2-72B & 69.0 & 79.4 & 80.3 & 66.5 & 15.6 & 42.4 & 66.4 \\
+ CoT & \textbf{78.8} & 85.7 & 77.9 & 79.5 & 30.3 & 39.8 & 64.0 \\
+ CoMAT \cite{leang2024comatchainmathematicallyannotated} & 72.4 & 83.9 & - & 81.7 & 32.2 & - & - \\
+ \QuaSAR & 77.5 & \textbf{86.2} & \textbf{84.3} & \textbf{83.5} & \textbf{36.2} & \textbf{48.2} & \textbf{69.0} \\ 
\bottomrule
\end{tabular}

\caption{Performance comparison using \QuaSAR as in-context learning strategy (\S \ref{sec:SiC_Application}) across multiple tasks and models (\S \ref{sec:exp_setup}). The results are obtained using zero-shot prompting as baselines, CoT \cite{kojima2022large}, CoMAT \cite{leang2024comatchainmathematicallyannotated} and Faithful CoT (FCoT) \cite{lyu2023faithful} as the main comparison.}
\label{tab:results_ICL}
\end{table*}

\section{Experiments}
\label{sec:exp_setup}
We evaluate \QuaSAR on complex mathematical problems, commonsense reasoning, and
natural language understanding tasks (\S \ref{sec:data}). We perform the evaluation phases by following standard approaches used to assess question-answering tasks (\S \ref{sec:evaluation}) on models presented in \S \ref{sec:models}.

\subsection{Tasks \& Datasets}
\label{sec:data}
We evaluate the operability of \QuaSAR on tasks involving complex reasoning and natural language inference. These tasks are best exemplified by the following categories:

\paragraph{Symbolic Tasks} We use GSM8K \cite{cobbe2021training}, SVAMP \cite{patel-etal-2021-nlp}, AQuA \cite{ling2017program}, MMLU-Redux \cite{gema2024we} and Olympiad Bench \cite{he2024olympiadbench} covering various mathematical topics, including abstract algebra, elementary, college-level and high-school mathematics.  
These datasets include multiple-choice questions (AQUA, MMLU-Redux) and math-world problems (GSM8K, MSVAMP, Olympiad Bench).

\paragraph{Natural Language Tasks}
We use Graduate-Level Google-Proof Q\&A Benchmark (GPQA) \cite{rein2023gpqagraduatelevelgoogleproofqa} and Reading Comprehension Benchmark Requiring Discrete Reasoning Over Paragraphs (DROP) \cite{dua-etal-2019-drop}. GPQA presents complex, open-ended questions that resist specific searches and require models to synthesise knowledge across multiple sources or reason critically to generate answers. DROP focuses on questions requiring discrete reasoning, such as arithmetic operations, logical comparisons, or event tracking, requiring the model to extract and manipulate information from a given passage. 

\subsection{Evaluation Metrics}
\label{sec:evaluation}
We used exact-match for the multiple-choice question-answering task, requiring the predicted answer to match the correct one. This guarantees evaluation based on complete responses, addressing clarity concerns in tasks like MMLU-Redux. For string-matching answers, we used exact matches in GSM8K. Moreover, to have a comprehensive evaluation, we use GPT-4o-mini as a benchmark to evaluate how well the model's answers aligned with the ground truth. Details are described in Appendix \ref{app:LLM_evaluator}.

\subsection{Models} 
\label{sec:models}

Experiments were performed on GPT-4o \cite{achiam2023gpt}, Qwen2 \cite{yang2024qwen2} and Llama-3 \cite{grattafiori2024llama3herdmodels}. While we selected the first two models to allow for a detailed comparison with related work and CoT frameworks, Llama-3  was chosen for its adaptability and the presence of releases with a small number of parameters that allow for additional tuning steps. 

\paragraph{Baselines} We compared \QuaSAR against two baselines using the same greedy decoding strategy, fixing the temperature to 0. The baselines include: (1) standard zero-shot prompting, (2) CoT prompting \cite{kojima2022large}. 
Moreover, we include Faithful CoT \cite{lyu2023faithful}, FLAIRE \cite{arakelyan2024flarefaithfullogicaidedreasoning}, and CoMAT \cite{leang2024comatchainmathematicallyannotated} for additional comparison. 

\paragraph{\QuaSAR Application} We use \QuaSAR as ICL and for generating tuning demonstrations. In both configurations, we instruct the models via the prompt in Appendix \ref{app:SiC_prompt}. We conduct instruction-tuning of the models using the demonstrations described in Appendix \ref{app:annotation} and the configurations in Appendix \ref{app:tuning_configuration}.

\begin{table*}[h]
\centering
\resizebox{\textwidth}{!}{
\begin{tabular}{lccccccc}
\toprule
\multirow{2}{*}{\textbf{Model}} & \multicolumn{5}{c}{\textbf{Symbolic}} & \multicolumn{2}{c}{\textbf{Natural Language}} \\
\cmidrule(lr){2-6} \cmidrule(lr){7-8}
& \textbf{AQuA} & \textbf{GSM8K} & \textbf{SVAMP} & \textbf{MMLU-Redux} & \textbf{OlyBench} & \textbf{GPQA} & \textbf{DROP} \\
\midrule
Llama-3-8B & 65.2\bcol{67.3} & 73.8\bcol{79.9} & 70.0\bcol{73.8} & 60.2\bcol{63.0} & 10.9\bcol{13.2} & 32.8\bcol{33.7} & 58.4\bcol{60.2} \\

+ CoT & 69.6\bcol{72.2} & 80.4\bcol{82.6} & 76.3\bcol{78.8} & 64.5\bcol{65.9} & 12.4\bcol{14.7} & 34.0\bcol{35.2} & 57.9\bcol{59.3} \\
+ FLARE \cite{arakelyan2024flarefaithfullogicaidedreasoning} & 62.9 & 72.4 & 86.0 & - & - & - & - \\
+ \QuaSAR  & 67.2\gcol{78.4} & 77.2\gcol{83.0}  & 77.3\gcol{82.6}  & 63.0\gcol{67.2}  & 13.0\gcol{16.6}  & 33.1\gcol{39.2}   & 58.7\gcol{63.9}  \\

\midrule
Llama-3-1B & 39.2\bcol{40.3} & 44.8\bcol{45.8} & 49.5\bcol{50.8} & 28.3\bcol{30.1} & 6.5\bcol{7.1} & 25.4\bcol{26.9} & 52.5\bcol{53.0} \\

+ CoT & 50.7\bcol{52.0} & 59.3\bcol{60.9} & 58.2\bcol{59.9} & 34.0\bcol{34.7} & 8.2\bcol{10.6} & 27.6\bcol{28.7} & 54.4\bcol{55.0} \\

+ \QuaSAR & 51.6\gcol{55.4} & 58.1\gcol{62.8} & 60.4\gcol{64.5} & 34.2\gcol{40.0} & 9.8\gcol{14.6} & 26.6\gcol{29.4} & 54.1\gcol{57.2} \\

\midrule

Qwen2-7B & 62.9\bcol{63.7} & 70.4\bcol{71.6} & 66.9\bcol{67.2} & 65.5\bcol{66.3} & 10.5\bcol{10.9} & 32.0\bcol{32.7} & 55.3\bcol{54.2} \\

+ CoT & 79.1\bcol{80.3} & 82.8\bcol{83.6} & 73.2\bcol{74.9} & 79.2\bcol{80.0} & 9.8\bcol{10.7} & 33.7\bcol{34.0} & 56.0\bcol{56.8} \\

+ CoMAT \cite{leang2024comatchainmathematicallyannotated} & 72.4 & 83.9 & - & 79.8 & 32.2 & - & - \\

+ \QuaSAR & 72.6\gcol{78.3} & 81.7\gcol{85.6} & 69.2\gcol{75.0} & 75.9\gcol{80.3} & 27.8\gcol{36.5} & 29.5\gcol{35.2} & 54.6\gcol{60.0} \\ 

\midrule

Qwen2-1.5B & 56.8\bcol{57.2} & 61.4\bcol{62.0} & 59.2\bcol{60.0} & 41.7\bcol{42.4} & 6.9\bcol{7.4} & 21.4\bcol{21.9} & 49.8\bcol{50.8} \\

+ CoT & 58.7\bcol{59.9} & 64.7\bcol{65.8} & 63.6\bcol{65.0} & 46.3\bcol{47.8} & 7.8\bcol{9.1} & 25.4\bcol{26.9} & 51.2\bcol{52.5} \\

+ \QuaSAR & 57.6\gcol{62.2} & 64.2\gcol{69.8} & 65.4\gcol{70.2} & 44.8\gcol{49.5} & 8.2\gcol{11.8} & 26.6\gcol{31.0} & 50.8\gcol{57.3} \\

\bottomrule
\end{tabular}
}
\caption{Performance comparison using \QuaSAR, CoT \cite{kojima2022large}, FLARE \cite{arakelyan2024flarefaithfullogicaidedreasoning} and, CoMAT \cite{leang2024comatchainmathematicallyannotated}as in-context learning strategies. Moreover, we report in brackets the performances obtained using these strategies as annotation approaches for tuning models (complete table in Appendix \ref{app:results_tuning_SiC_complete}).}
\label{tab:results_tuning_SiC}
\end{table*}

\section{Results \& Discussions}
The results in Tables \ref{tab:results_ICL} and \ref{tab:results_tuning_SiC} compare \QuaSAR with baselines, CoT and related work across various tasks. 
\QuaSAR outperforms CoT in most tasks requiring advanced mathematical reasoning (Symbolic task), reading comprehension and logical reasoning (Natural Language task).
In particular, two different results emerge in the application of \QuaSAR: when it is employed as in-context learning strategy in higher-scale models, it consistently outperforms other strategies; when \QuaSAR is employed in smaller-scale models, it does not obtain the same benefits, as discussed in \S \ref{sec:SiC_in-context}. 
On the other hand, when \QuaSAR is used as an annotation strategy for delivering demonstrations operated to refine smaller-scale models, the performances are significantly higher compared to the models instructed via standard CoT demonstrations, as examined in \S \ref{sec:SiC_as_annotator}. 

Overall, our experiments demonstrate the benefit of quasi-symbolic abstractions for complex reasoning tasks, and provide evidence of improved robustness on challenging adversarial variations (\S \ref{sec:SiC_impact}).

\subsection{\QuaSAR as In-Context Learning Strategy}
\label{sec:SiC_in-context}
Table \ref{tab:results_ICL} reports the results of \QuaSAR when adopted as an In-Context Learning (ICL) strategy. We observe general robust improvement over the baseline models (with an improvement of 19.1\% for GPT-4o, 11.8\% for Llama-3-70B and 17.2\% for Qwen2-72B); the results show that the role of \QuaSAR as ICL is foremost noticeable for higher-scale LLMs. \QuaSAR consistently outperforms CoT, Faithful CoT and CoMAT. \QuaSAR also delivers overall improvements on smaller-scale models. Table \ref{tab:results_tuning_SiC} shows an improvement over the baseline of 5.2\% for Llama-3-8B, 13.4\% for Llama-3-1B, 10.5\% for Qwen2-7B and  8.3\% for Qwen2-1.5B. However, comparing \QuaSAR to CoT on smaller models, we observe a decrease in performance, indicating that such models fail to follow the quasi-symbolic reasoning process induced by \QuaSAR.

\subsection{\QuaSAR as Annotation Strategy}
\label{sec:SiC_as_annotator}

Table \ref{tab:results_tuning_SiC} (values between the brackets and detailed in Appendix \ref{app:results_tuning_SiC_complete}) reports the results of \QuaSAR when adopted as annotation strategy for different models. From the results, it clearly emerges that \QuaSAR is consistently effective in enhancing the performance of Llama and Qwen2 models when used to generate reasoning demonstrations via GPT-4o. In particular, we found that \QuaSAR outperforms other tuning approaches, including baseline SFT on target answers and SFT on demonstrations delivered via CoT. 

\subsection{The impact of \QuaSAR}
\label{sec:SiC_impact}

The step-wise reasoning chain generations elicited by \QuaSAR have an optimal impact on the downstream performances when \QuaSAR is used as ICL and for generating demonstrations.

\begin{table}[t]
\small
\centering
\begin{tabularx}{\columnwidth}{p{1.5cm}XXXXXX}
    \textbf{Task} & \rotatebox{90}{\textbf{w/o\textit{(1)}}} & \rotatebox{90}{\textbf{w/o\textit{(2)}}} & \rotatebox{90}{\textbf{w/o\textit{(3)}}} & \rotatebox{90}{\textbf{w/o\textit{(4)}}} & \rotatebox{90}{\textbf{w/o\textit{(1-2)}}} & \rotatebox{90}{\textbf{w/o\textit{(3-4)}}} \\
    \toprule    
    \textbf{AQuA}   & \textbf{-1.9} & \textbf{-3.6} & \textbf{-3.7} & \textbf{-2.9} & \textbf{-3.7} & -2.7 \\
    \textbf{GSM8K}  & \textbf{-2.1} & \textbf{-4.2} & \textbf{-3.9} & -1.3 & \textbf{-3.5} & -2.6 \\
    \textbf{MMLU-R}  & -0.7 & -3.0 & -3.2 & \textbf{-3.2} & -2.2 & \textbf{-2.8} \\
    \textbf{OlyBench}  & \textbf{-1.9} & -3.1 & \textbf{-3.7} & -2.1 & \textbf{-3.8} & -2.3 \\
    \textbf{GPTQ}  & -1.6 & \textbf{-3.9} & \textbf{-3.6} & \textbf{-3.2} & \textbf{-4.1} & \textbf{-2.9}\\
    \hdashline
    \textbf{Avg}       & -1.8 & -3.5 & -3.4 & -2.5 & -3.2 & -2.8 \\
    \bottomrule
\end{tabularx}
\caption{Performance change without (w/o) \QuaSAR step obtained from GPT-4o. *(Bold values over the average).}
\label{tab:ICL_missing_step}
\end{table}

\paragraph{Step-wise roles for ICL}
Table \ref{tab:ICL_missing_step} displays the difference in accuracy compared to \QuaSAR with all steps. We show that each step impacts \QuaSAR's operability. In particular, eliminating step 1 (i.e., \textit{w/o(1)}) affects the final accuracies (-1.8 on average). This suggests that the initial abstraction step is important for final performance but is not decisive, especially in tasks such as GPTQ and MMLU-Redux. In contrast, steps 2 (i.e., formalisation) and 3 (i.e., explanation), play a crucial role, indeed, a stable drop of more than 3.5 points can be observed. In this case, the tasks that suffer the most are the mathematical subset (AQuA and GSM8K). Step 4, reserved for the strict generation of the final answer, is more decisive in multiple-choice than in mathematical tasks. Finally, by eliminating pairs of steps (i.e. \textit{w/o(1-2)} and \textit{w/o(3-4)}), it can be seen that there are significant drops in both mathematical tasks (see GSM8K, AQuA and OlympicBench) and language-related reasoning tasks (see MMLU-Redux and GPTQ). The combination of the four steps from these results positively impacts reasoning capabilities, and they all contribute significantly to the final performances.

\begin{figure}[h]
    \centering
    \includegraphics[width=\linewidth]{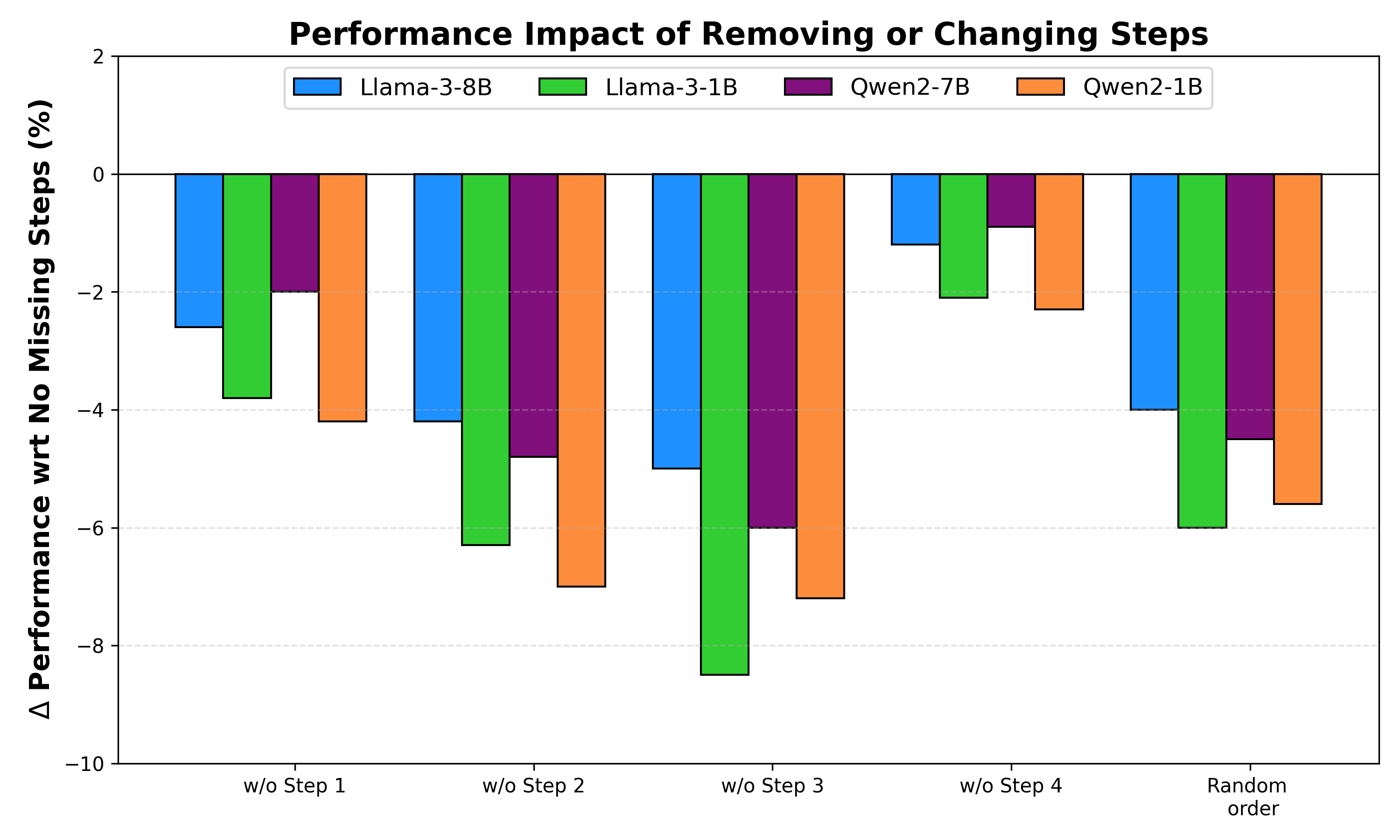}
    \caption{Performance differences $(\Delta)$ for each tuned model. We analyse the impact of each component on tuning by eliminating (w/o) or random shuffling the four \QuaSAR steps.
    }
\label{fig:missing_steps_performances}
\end{figure}

\paragraph{Step-wise role for Annotations}
Figure \ref{fig:missing_steps_performances} displays the difference in accuracy using entire \QuaSAR for generating demonstrations. As in the case of ICL, the steps in the demonstrations have specific importance for instructing models. Indeed, it can be observed that the instructed models perform worse by eliminating central steps such as Step 2 and Step 3. In contrast, removing step 4 duplicated to the response has moderate adverse effects (performance drop of no more than two points). Finally, it can be seen that the order in which the steps are delivered in the demonstrations also has a positive impact. Delivering the demonstrations randomly shuffled negatively impacts performances, dropping around 4 average points.

\begin{table}[t]
\small
\centering
\setlength{\tabcolsep}{4pt}
\begin{tabularx}{0.48\textwidth}{c|lccc}
    \toprule
    \textbf{} & \textbf{Task} & \textbf{Baseline} & \textbf{CoT} & \textbf{\QuaSAR}  \\
    \midrule
    \multirow{4}{*}{\rotatebox{90}{\textbf{GPT-4o}}} 
    & MMLU-Redux        & 79.8 & 88.2 & \textbf{90.3} \\
    & \textit{-choices shuffled}     & 78.6\bcol{-1.2} & 86.8\bcol{-1.2} & \textbf{90.3}\gcol{0.0}  \\ 
    & GSM-Symbolic & 94.0 & 95.5 &  96.5 \\
    & \textit{-2nd choice} & 89.7\bcol{-4.3} & 90.8\bcol{-4.7} &  95.3\gcol{-1.2} \\
    \midrule
    \multirow{4}{*}{\rotatebox{90}{\tiny \textbf{Llama3-70B}}} 
    & MMLU-Redux        & 70.8 & 81.9 & 85.0 \\
    & \textit{-choices shuffled}     & 68.7\bcol{-0.9} & 81.0\bcol{-0.9} & \textbf{84.8}\gcol{-0.2}  \\ 
    & GSM-Symbolic & 84.2 & 85.3 &  87.8 \\
    & \textit{-2nd choice} & 82.6 \bcol{-1.6} & 83.7\bcol{-1.6} &  87.2\gcol{-0.4} \\
    \midrule
    \multirow{4}{*}{\rotatebox{90}{ \tiny \textbf{Llama3-8B}}} 
    & MMLU-Redux   & 30.2 & 31.6 & 37.6 \\
    & \textit{-choices shuffled}   & 27.0\bcol{-3.2} & 30.4\bcol{-1.2} & 37.3\gcol{-0.3}  \\ 
    & GSM-Symbolic & 46.7 & 60.2 &  65.3 \\
    & \textit{-2nd choice} & 44.9\bcol{-1.8} & 58.4\bcol{-1.8} &  64.8\gcol{-0.5} \\
    \bottomrule
\end{tabularx}
\caption{Performance obtained by changing the order of choices randomly (MMLU-Redux) and using a perturbed version of mathematical tasks (GSM-Symbolic). *(accuracies differences in brackets) }
\label{tab:ablation_SiC_ICL}
\end{table}

\subsection{Robustness \& Ablation Analysis}

\paragraph{In-context Robustness}
To assess the robustness of \QuaSAR as an ICL strategy, we evaluated two different phenomena: \textit{(i)} the order swapping of choices in MMLU-Redux \cite{gema2024we} as proposed by \citet{leang2024comatchainmathematicallyannotated} and \textit{(ii)} the performance on a more complex version of GSM8K designed to test robustness to superficial variations (i.e., GSM-Symbolic \cite{mirzadeh2024gsmsymbolicunderstandinglimitationsmathematical}). Table \ref{tab:ablation_SiC_ICL} shows that \QuaSAR consistently achieves the same performances with considerably less variation than CoT. This indicates the positive impact of quasi-symbolic abstractions on the robustness of the models.

\begin{figure}[t]
    \centering    \includegraphics[width=0.95\linewidth]{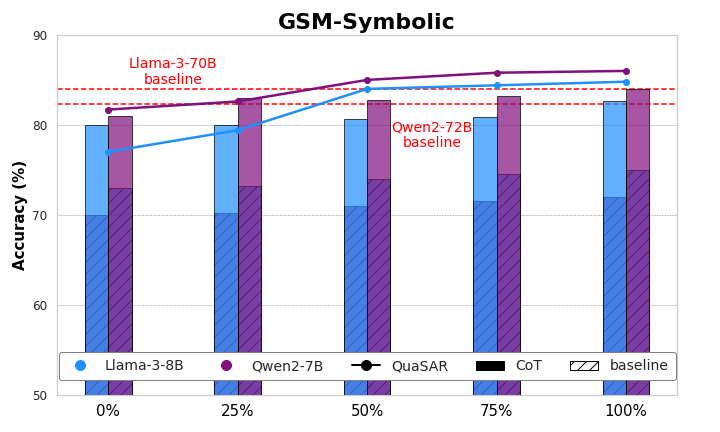}
    \caption{Performances using \QuaSAR as demonstration tuning by scaling training data. We replicated experimental settings proposed in \S \ref{sec:exp_setup}, changing the number of tuning instructions. *Appendix \ref{app:scaling_data_additional} reports additional evaluaitons.}
    \label{fig:scaling_training_data}
\end{figure}

\paragraph{Training Efficiency}
Figure \ref{fig:scaling_training_data} shows the performance of tuned models on GSM-Symbolic \cite{mirzadeh2024gsmsymbolicunderstandinglimitationsmathematical} using \QuaSAR, CoT, and baseline demonstrations as the number of training examples increases. While the number of demonstrations in \QuaSAR plays a significant role in shaping the final performance, our findings reveal that models trained with \QuaSAR demonstrations surpass those trained with CoT demonstrations consistently. Moreover, it is possible to observe that models instructed via \QuaSAR demonstrations outperform their respective models with more parameters with 50\% of the training examples in the case of Llama-3-8B and 25\% in the case of Qwen2-7B. Notably, \QuaSAR consistently outperforms other supervised training approaches even with this reduced dataset size, emphasising the superior quality of the training signal provided by \QuaSAR demonstrations.

\paragraph{Additional Analysis}

Finally, we produced further analyses to investigate the error flow, the degree of self-correction and the transferability of the approach. In the first analysis, we showed the error rate of our \QuaSAR. In Appendix \ref{app:error_propagation}, we present a particular analysis to investigate the error rate of each step, arguing for its validity.

In the second analysis, we studied the self-correction capability of the models trained and tuned via \QuaSAR and via CoT. Appedinx \ref{app:self_consistency} reports the results of the self-assessment on incorrect generations, showing that the outputs generated via \QuaSAR are actually easier to correct, as the output rationale, being structured, is simpler to correct.

In the final analysis, reported in Appendix \ref{app:add_evaluations}, we showed the elasticity and adaptability of \QuaSAR in tasks different from those proposed in the main analysis, confirming the results obtained and discussed in the previous sections.

\section{Related Work}

\paragraph{Logical Reasoning} Logical reasoning tasks require the capability to process complex logical structures \cite{cummins1991conditional}. Traditional methodologies contain rule-based systems \cite{robinson1965machine} and neural network-based paradigms \cite{amayuelas2022neural,gerasimova2023comparative} for solving and manipulating symbolic representations. Recent advancements introduced hybrid frameworks \cite{pan2023logic,ye2024satlm,jiang2024leanreasoner}, which integrate large language models (LLMs) into symbolic reasoning pipelines \cite{quan2024verificationrefinementnaturallanguage}. These frameworks operate LLMs to map natural language inputs into symbolic syntax, subsequently processed by external reasoning tools. This integration improves reasoning performance through techniques such as self-consistency \cite{wang2023plan,zhang2022automatic}. Nevertheless, these frameworks commonly depend on external tools predicated on the assumption that LLMs lack the reliability to parse symbolic expressions with the precision of rule-based reasoning systems alone.

\paragraph{Symbolic Reasoning}
Symbolic reasoning integrates natural language (NL) and symbolic language (SL) to decompose complex queries into sub-problems solved by SL programs and deterministic solvers, ensuring interpretability and precision \cite{lyu2023faithful}. Recent efforts have leveraged LLMs to decrease dependence on SL programs \cite{xu2024faithful}, but these approaches primarily address logical reasoning and depend on verifiers for accuracy, limiting their applicability to complex mathematical tasks.

On the other side, Chain-of-Thought (CoT) strategies have demonstrated significant performance improvements in mathematical symbolic reasoning \cite{jiang2024llms}, reinforced by advancements in problem understanding \cite{zhong2024achieving}, structured formats \cite{tam2024let}, and supervision models \cite{ranaldi2024aligning,ranaldi-freitas-2024-self,jiang2024rationalystpretrainingprocesssupervisionimproving}. Further, premise selection and symbolic frameworks have facilitated systematic evaluations across logical and mathematical reasoning \cite{meadows2023symbolic,ferreira2020premise}.

\section{Future Works}
In future developments, we plan to extend our contribution to non-English languages to broaden the beneficial impacts and operability of reasoning for multilingual alignment. To this end, we will use our approach in the multilingual extension of GSM-Symbolic \cite{mirzadeh2024gsmsymbolicunderstandinglimitationsmathematical} proposed by \citet{ranaldi-pucci-2025-multilingual}. Furthermore, we would like to investigate the extent to which our framework can be applied in scenarios where retrieval-augmented LLMs approaches are used, such as our parallel contributions, where we propose techniques to resolve knowledge conflicts in retrieved documents \cite{ranaldi2025improvingmultilingualretrievalaugmentedlanguage,ranaldi-etal-2025-eliciting}.

\section{Conclusion}

Complex reasoning tasks often require the co-existence of natural language and symbolic abstractions. Many existing methods based on CoT struggle to ensure consistency and robustness, particularly when handling tasks with shuffled answer options or superficial lexical variations.
In this paper, we proposed Quasi-Symbolic Abstract Reasoning (\QuaSAR) to address these challenges. This simple yet powerful framework enables LLMs to tackle such tasks by breaking them down into systematic, quasi-symbolic step-by-step reasoning. By employing \QuaSAR as an in-context learning strategy and a tool for constructing demonstrations, we improved the performance of smaller models and provided a comprehensive analysis across diverse benchmarks.
Our experiments demonstrate that \QuaSAR surpasses traditional CoT reasoning methods by delivering transparent and consistent reasoning trajectories. \QuaSAR excels across tasks of varying complexity, achieving state-of-the-art performance and improving robustness. \QuaSAR delivers a scalable and effective solution for complex reasoning, enhancing faithfulness, verifiability, and reliability while outperforming conventional Chain-of-Thought approaches.

\section*{Acknowledgements}
This work was funded by the Swiss National Science Foundation (SNSF) project ``NeuMath'' (200021\_204617),  Innosuisse project ``SINFONIA'' (n. 104.170 IP-ICT), by the CRUK National Biomarker Centre, and supported by the Manchester Experimental Cancer Medicine Centre, the NIHR Manchester Biomedical Research Centre and UK Research and Innovation under the UK government’s Horizon Europe funding guarantee grant number 10039436. 

\bibliography{custom}

\appendix

\begin{table*}
\section{\QuaSAR Prompting Template}
\label{app:SiC_prompt}

\begin{small}
\begin{tcolorbox}[colback=lightgray, colframe=customisedblue, sharp corners=south, rounded corners=north]

\begin{tcolorbox}[colback=white, colframe=customisedblue, rounded corners=south, rounded corners=north]
\textbf{\#Role} \\
You are an experienced expert skilled in answering complex problems through logical reasoning and structured analysis. 
\end{tcolorbox}

\begin{tcolorbox}[colback=white, colframe=customisedblue, rounded corners=south, rounded corners=north]
\textbf{\#Task} \\
You are presented with a problem that requires logical reasoning and systematic problem-solving. Please answer the question following these steps rigorously.
\end{tcolorbox}

\begin{tcolorbox}[colback=white, colframe=customisedblue, rounded corners=south, rounded corners=north]
\textbf{\#Steps} 

\begin{adjustwidth}{0.5cm}{0cm} 

    \textbf{1)} Please consider the following question and exemplify the relevant predicates, variables, and constants. Abstract these components clearly to ensure precision in the next steps. \textit{Do not omit any details and strive for maximum precision in your explanations. Refer to this step as \textcolor{customisedblue}{Abstraction ($s_1$)}} \\

    \textbf{2)} For each predicate, variable and constant defined in \textcolor{customisedblue}{$s_1$}, translate the question in formal symbolic representation. Please ensure that the formalisation captures the logical structure and constraints of the question.
    \textit{For clarity, provide the exact formalisation of each component exemplified in \textcolor{customisedblue}{$s_1$}, referencing their corresponding definitions. Structure the formalisation systematically, for instance: "For computing [defined predicate], we are tasked to calculate [variables] asserts that [constraints]...". Refer to this step as \textcolor{customisedblue}{Formalisation ($s_2$)}}  \\ 

    \textbf{3)} Please consider the formalisation in \textcolor{customisedblue}{$s_2$} in detail, ensure this is correct and solve the question by breaking down the steps operating a symbolic representation. Combine variables, constants, and logical rules systematically at each step to find the solution.
    \textit{For clarity, provide clear reasoning for each step. Structure the explanation systematically, for instance: "Step 1: Calculate... Step 2:....". Refer to this step as \textcolor{customisedblue}{Explaination ($s_3$)}} \\
    
    \textbf{4)} In conclusion, behind explaining the steps supporting the final answer to facilitate the final evaluation, extract the answer in a short and concise format by marking it as “\textbf{The answer is }” \textit{At this stage be strict and concise and refer to this step as \textcolor{customisedblue}{Answering ($s_4$).}}
\end{adjustwidth}
    
\end{tcolorbox}

\begin{tcolorbox}[colback=white, colframe=customisedblue, rounded corners=south, rounded corners=north]
\textbf{\#Question} \\
\{\texttt{\textbf{question}}\}
\end{tcolorbox}

\end{tcolorbox}
\end{small}
\caption{The Step-wise Instruction Chain (\QuaSAR) framework instructs the model to deliver step-wise reasoning paths that lead the models to solve the task by delivering a formalised strict final answer.}
\label{tab:prompt_SiC}

\end{table*}

\begin{table}[]
\section{Annotations Pipeline}
\label{app:annotation}
As introduced in \S \ref{sec:methods}, we use our Step-wise Instruction Chain (\QuaSAR) to lead Llama-3-1B, -8B, Qwen2-7B and 1B in solving complex tasks by breaking down the solution using the \textit{reasoning process} described in \S \ref{sec:tuning}. Since \QuaSAR alone does not fully leverage the capabilities of the baseline models—significantly smaller models without further tuning, as shown in Table \ref{tab:results_tuning_SiC}—we use GPT-4o (GPT-4) as an annotation model. GPT-4 is systematically prompted using the instructions detailed in Appendix \ref{app:SiC_prompt}.

GPT-4 is used to generate synthetic demonstrations to train models in delivering \QuaSAR's step-wise reasoning methods. However, while GPT-4 follows the instructions exhaustively, its outputs may include errors or misleading information. To address this, we evaluated the quality of the generated demonstrations, filtering out inaccurate examples to refine the instruction set. Specifically, we removed all incorrect answers (i.e., outputs that do not match the exact target string metric, referred to as \textit{exact-match}). Finally, we verified that all essential steps were correctly encoded in the remaining demonstrations using GPT-4o-mini and the prompt in Appendix \ref{tab:prompt_answer_evaluation}
\end{table}

\begin{table}[]
\section{Evaluation Metrics}
\label{app:LLM_evaluator}

We used a double-check to assess the accuracy of the responses delivered in the different experiments. In the first step, we used an exact-match heuristic (this was used for most of the evaluations, especially in cases of multiple-choice QA). However, since some experiments required a more accurate response check, we used GPT-4o as a judge. Hence, we prompt the model as follows:

\begin{tcolorbox}[
    colback=lightgray,
    colframe=gray!75!black,
    colbacktitle=gray!90!white,
    fonttitle=\bfseries,
    width=\columnwidth,
    boxrule=0.5pt,
    arc=4pt,
    auto outer arc,
]
\small
\textbf{\#Role:} \\
You are an experienced expert skilled in answering complex problems through logical reasoning and structured
analysis. \\
\textbf{\#Task:} \\
Given the following "\#Sentences", you are a decider that decides whether the "Generated Answer" is the same as the "Target Answer". 
If the output doesn't align with the correct answer, respond with '0', whereas if it's correct, then respond with '1'. \textit{Please, do not provide any other answer beyond `0' or `1'.} \\
\textbf{\#Senteces:} \\
Generated Answer: \{model\_result\} \\
Target Answer: \{correct\_answer\}. \\
\end{tcolorbox}
\label{tab:prompt_answer_evaluation}
\end{table}

\begin{table}[h]
\section{Data Composition}
\label{app:data_composition}

We evaluated \QuaSAR using the tasks introduced in \S \ref{sec:data}. Although these tasks are most often used to assess the performance of LLMs, they often do not have dedicated sets for evaluation and training. Therefore, to use \QuaSAR both as an in-context prompting approach and as an instruction generation approach, we divided the datasets into training and testing. Table \ref{tab:data_composition} shows the instances of each dataset in training and testing. Where we did not find split data already, we produced a splitting, which is also displayed in Table \ref{tab:data_composition}.

\vspace{0.5cm}

\small
\centering
\setlength{\tabcolsep}{3.5pt}
\begin{tabularx}{0.46\textwidth}{lcccc}
    \toprule
    \textbf{Task} & \textbf{Total} & \textbf{Test} & \textbf{Trainig Set} & \textbf{Testing Set} \\
    \midrule
    \textbf{AQuA}        & $254$ & $254$ & Yes & $254$ \\
    \textbf{GSM8K}     & $8,02k$ & $1,32k$ & Yes  & $1,32k$ \\
    \textbf{SVAMP}      & $700$ & $700$ & Yes   & $700$ \\
    \textbf{MMLU-Redux}      & $1k$ & $1k$ & No & $1k$ \\
    \textbf{OlyBench}      & $2,5k$ & $1,5k$ & Yes   & $500$ \\
    \textbf{GPQA}      & $198$ & - & No  & $198$ \\
    \textbf{DROP}      & $2,5k$ & $1,5k$ & Yes & $500$ \\
    \bottomrule
\end{tabularx}
\caption{Data used to evaluate \QuaSAR as in-context learning approach. When training set are present we tagged as "Yes". *($1k$ is equal to $1000$).}
\label{tab:data_composition_test_data}

\vspace{0.5cm}

\small
\centering
\setlength{\tabcolsep}{3.5pt}
\begin{tabularx}{0.4\textwidth}{lcccc}
    \toprule
    \textbf{Task} & \textbf{Total} & \textbf{Correct} &  & \textbf{Used} \\
    \midrule
    \textbf{AQuA}        & $97k$ & $3.0k$ &  & $1.0k$ \\
    \textbf{GSM8K}     & $6k$ & $2.04k$ &  & $0.8k$ \\
    \textbf{OlyBench}     & $420$ & $250$ &  & $250$ \\
    \textbf{DROP}      & $7,5k$ & $1k$ &  & $350$ \\
    \hdashline
    \textbf{Total}     & $22k$   & $6,9k$ &  & \textbf{\textit{2,4k}} \\
    \bottomrule
\end{tabularx}
\caption{Data used to construct \QuaSAR demonstrations. We applied the annotation (\S \ref{sec:annotation_strategy}) and obtained the following answers, filtered according to the heuristics in Appendix \ref{app:annotation}, and balanced for the tasks. }
\label{tab:data_composition}
\end{table}

\begin{table}[]
\section{Additional Task}
\label{app:add_evaluations}

\begin{small}
\begin{center}
  \begin{tabularx}{0.46\textwidth}{p{1.8cm}<{\centering}p{1cm}<{\centering}p{1cm}<{\centering}p{1cm}<{\centering}}
    \toprule
    \multirow{1}{*}{\textbf{Method}} & \multicolumn{1}{c}{\textbf{MATH}} & \multicolumn{1}{c}{\textbf{XCOPA}} & \multicolumn{1}{c}{\textbf{MGSM}} \\
    \midrule
    baseline & 70.4 & 84.4 & 90.5 \\
    CoT      & 76.8 & 88.6 & 91.0 \\
    \QuaSAR   & \textbf{79.5} & \textbf{89.2} & \textbf{93.4} \\
    \bottomrule
  \end{tabularx}
\end{center}
\end{small}
\caption{GPT-4o performances on MATH, XCOPA, and MGSM.}
\label{tab:quasar_gpt4o}

\begin{small}
\begin{center}
  \begin{tabularx}{0.46\textwidth}{p{1.8cm}<{\centering}p{1cm}<{\centering}p{1cm}<{\centering}p{1cm}<{\centering}}
    \toprule
    \multirow{1}{*}{\textbf{Method}} & \multicolumn{1}{c}{\textbf{MATH}} & \multicolumn{1}{c}{\textbf{XCOPA}} & \multicolumn{1}{c}{\textbf{MGSM}} \\
    \midrule
    baseline & 30.0 & 56.4  & 59.0 \\
    CoT      & 33.0 & 56.9   & 60.8 \\
    \QuaSAR   & \textbf{36.4} & \textbf{65.0}   & \textbf{66.9} \\
    \bottomrule
  \end{tabularx}
\end{center}
\end{small}
\caption{Llama-3-8B performances on MATH, XCOPA, and MGSM.}
\label{tab:quasar_gpt4o}
\end{table}

\begin{table}[]
\section{Training Setup}
\label{app:tuning_configuration}
To evaluate the impact of \QuaSAR demonstrations on smaller models (\S \ref{sec:methods}), we use the annotations produced following the \QuaSAR strategy (\S \ref{sec:annotation_strategy}). For a fair comparison, we generated CoT annotations and naive output without any prompting approach using GPT-4 on the same instances. Then, we train selected models using \QuaSAR, CoT and standard output demonstrations. We fine-tuned the Llama-3 models for 3 epochs with a batch size of 32 and a learning rate equal to 3e-5 with a 0.001 weight decay and the Qwen2 models for the same epochs and batch size. Instead, a learning rate equal to 2e-5 with a 0.002 weight decay was used. 

\vspace{1cm}

\section{Models Vesions}
\label{app:model_versions}
\small
\centering 
\begin{tabular}{p{2.cm}|p{4.6cm}}
\textbf{Model} & \textbf{Version}  \\ 
\toprule
Llama-3-70B   &  meta-llama/Meta-Llama-3-70B-Instruct \\
Llama-3.1-8B   &  meta-llama/Meta-Llama-3-8B-Instruct \\
Llama-3.2-1B   &  meta-llama/Llama-3.2-1B-Instruct \\ 
\hline
Qwen2-72B   & Qwen/Qwen2-72B-Instruct  \\
Qwen2-7B   & Qwen/Qwen2-7B-Instruct  \\
Qwen2.5-1.5B   & Qwen/Qwen2.5-1.5B-Instruct  \\
\hline
GPT-4-o & OpenAI API (gpt-4o-2024-08-06)  \\
GPT-4-o-mini & OpenAI API (gpt-4o-mini-2024-07-18)  \\
\hline

\end{tabular}

\caption{List of the versions of the models proposed in this work, which can be found on huggingface.co. We used the configurations described in Appendix \ref{app:model_info} in the repositories for each model *(access to the following models was verified on 12 Jan 2024).}
\label{tab:versions_models}
\end{table}

\begin{table}[]
\section{Evaluation Scaling training Data}
\label{app:scaling_data_additional}
    \centering
    \includegraphics[width=\linewidth]{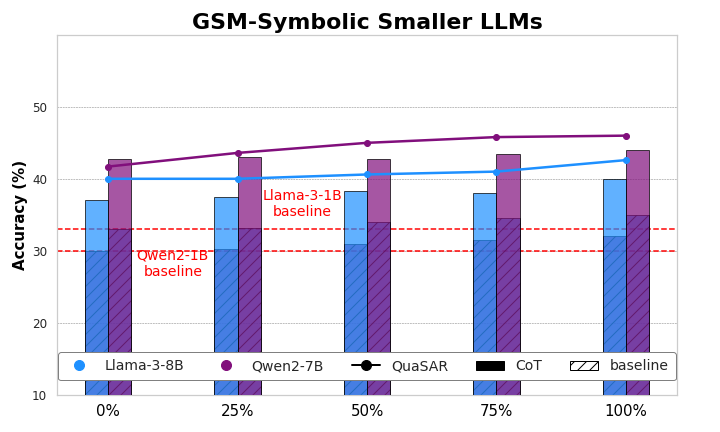}
    \caption{Performance assessment using \QuaSAR as demonstration tuning by scaling training data. We replicated experimental settings proposed in \S \ref{sec:exp_setup} changing the number of tuning instructions.}
    \label{fig:scaling_training_data_smaller}

\end{table}

\begin{table}[]
\section{Model and Hyperparameters}
\label{app:model_info}

As introduced in \S \ref{sec:models}, we propose different LLMs: (i) GPT-4o; (ii) three models from the Llama-3 family \cite{grattafiori2024llama3herdmodels}: Llama3-70B, Llama3.1-8B, Llama3.2-1B; (iii) three models of the Qwen2 family \cite{yang2024qwen2}: Qwen2-72B, Qwen2-7B and -1B.
GPT-4 is used via API, while for the others, we used versions detailed in Table \ref{tab:versions_models}.
As discussed in the limitations, our choices are related to reproducibility and the cost associated with non-open-source models.
The generation temperature used varies from $\tau = 0$ of GPT models to $\tau = 0.5$ of Llama models. We choose these temperatures for (mostly) deterministic outputs, with a maximum token length of 3500. The other parameters are left unchanged as recommended by the official resources. The code and the dataset will be publicly released upon acceptance of the paper.     
\end{table}

\begin{table}
\section{Error Propagation}
\label{app:error_propagation}

We provide a detailed analysis of error propagation across the four passages proposed in \S\ref {sec:methods}. We are quantifying the error rates attributed to each subcomponent, recognising that every stage performs a distinct function. The analysis was conducted on an ablation subset of GSM-Symbolic using GPT-4o. The error rate for each step was independently assessed via a manual verification process. The total failure rate across the full pipeline is \textbf{36\%}.

\vspace{1em}
\label{tab:error_propagation}
\begin{small}
\begin{center}
\begin{tabularx}{0.5\textwidth}{p{2.2cm}<{\raggedright}|p{3.1cm}<{\raggedright}r}
\toprule
\textbf{Stage} & \textbf{Description} & \textbf{Error Rate} \\
\midrule
Abstraction & Identification of relevant predicates, variables, constants & 8\% \\ \hdashline
Formalisation & Translation of input into symbolic/semi-symbolic structures & 12\% \\ \hdashline
Explanation & Step-by-step reasoning using the structured representation	 & 16\% \\ \hdashline
Answering & Generation of the final solution & 6\% \\ 
\bottomrule
\end{tabularx}
\end{center}
\end{small}
\vspace{1em}
\begin{small}
\begin{center}
\begin{tabularx}{0.5\textwidth}{p{2.5cm}<{\raggedright}p{2cm}<{\centering}p{2cm}<{\centering}}
\toprule 
\textbf{Step} & \textbf{Isolated Error} & \textbf{Cumulative Error} \\
\midrule
Abstraction & 8\% & 8\% \\ \hdashline
Formalisation & 12\% & 17\% \\ \hdashline
Explanation & 16\% & 32\% \\ \hdashline
Answering & 6\% & 36\% \\
\bottomrule
\end{tabularx}
\end{center}
\end{small}
\caption{Error rates per stage and cumulative error analysis. Each stage contributes independently and sequentially to the overall error rate.}
\end{table}

\begin{table*}[]
\section{Complete Results Smaller LLMs}
\label{app:results_tuning_SiC_complete}
\centering
\tiny
\begin{tabular}{lccccccc}
\toprule
\multirow{2}{*}{\textbf{Model}} & \multicolumn{5}{c}{\textbf{Symbolic}} & \multicolumn{2}{c}{\textbf{Natural Language}} \\
\cmidrule(lr){2-6} \cmidrule(lr){7-8}
& \textbf{AQuA} & \textbf{GSM8K} & \textbf{SVAMP} & \textbf{MMLU-Redux} & \textbf{OlyBench} & \textbf{GPQA} & \textbf{DROP} \\
\midrule
Llama-3-8B & 65.2 & 73.8 & 70.0 & 60.2 & 10.9 & 32.8 & 58.4 \\
Llama-3-1B$_{SFT}$ & 68.3 & 74.9 & 73.8 & 63.0 & 13.2 & 33.7 & 60.2 \\
+ CoT$_{ICL}$ & 69.6 & 80.4 & 76.3 & 64.5 & 12.4 & 34.0 & 57.9 \\
+ CoT$_{SFT}$ & 73.2 & 82.6 & 78.8 & 65.9 & 14.7 & 35.2 & 59.3 \\
+ FLARE \cite{arakelyan2024flarefaithfullogicaidedreasoning} & 62.9 & 72.4 & 86.0 & - & - & - & - \\
+ \QuaSAR$_{ICL}$  & 67.2 & 77.2  & 75.6  & 62.0  & 13.4  & 33.0   & 58.7  \\ 

+ \QuaSAR$_{SFT}$ & 74.8 &  \textbf{83.0} &  82.6 &  67.2 & 17.6 &  \textbf{39.2}  &  63.6 \\ 

+ \QuaSAR$_{SFT+ICL}$ & \textbf{75.2} &  82.8 &  84.7 &  \textbf{68.0} &  \textbf{17.8} &  \textbf{39.2}  &  \textbf{63.9} \\ 

\midrule
Llama-3-1B & 39.2 & 44.8 & 49.5 & 28.3 & 6.5 & 25.4 & 52.5 \\
Llama-3-1B$_{SFT}$ & 40.3 & 45.8 & 50.8 & 30.1 & 7.1 & 26.9 & 53.0 \\

+ CoT$_{ICL}$ & 50.7 & 59.3 & 58.2 & 34.0 & 8.2 & 27.6 & 54.4 \\

+ CoT$_{SFT}$ & 52.0 & 60.9 & 59.9 & 34.7 & 8.8 & 28.7 & 55.0 \\

+ \QuaSAR$_{ICL}$ & 51.6 & 58.1 & 60.4 & 30.2 & 10.6 & 26.6 & 54.1 \\ 

+ \QuaSAR$_{SFT}$ & 55.4 & \textbf{62.8} & 64.5 & 40.0 & 14.0 & \textbf{29.4} & 57.2 \\ 

+ \QuaSAR$_{SFT+ICL}$ & \textbf{56.0} & \textbf{62.8} & \textbf{64.9} & \textbf{40.8} & \textbf{14.6} & 29.3 & \textbf{57.7} \\ 

\midrule

Qwen2-7B & 62.9 & 70.4 & 66.9 & 65.5 & 10.5 & 32.0 & 55.3 \\
Qwen2-7B$_{SFT}$ & 63.7 & 71.6 & 67.2 & 66.3 & 10.9 & 32.7 & 56.2 \\

+ CoT$_{ICL}$ & 79.1 & 82.8 & 73.2 & 79.2 & 9.8 & 33.7 & 56.0 \\

+ CoT$_{SFT}$ & 80.3 & 83.6 & 74.9 & 80.0 & 11.7 & 35.0 & 56.8 \\

+ CoMAT \cite{leang2024comatchainmathematicallyannotated} & 72.4 & 83.9 & - & 79.8 & 32.2 & - & - \\

+ \QuaSAR$_{ICL}$ & 72.6 & 81.7 & 69.2 & 75.9 & 27.8 & 29.5 & 54.6 \\ 

+ \QuaSAR$_{SFT}$ & 
78.3 & \textbf{85.6} & 75.0 & 80.3 & \textbf{35.6} & 35.2 & \textbf{60.0} \\ 

+ \QuaSAR$_{SFT+ICL}$ & 
\textbf{79.0} & \textbf{85.6} & \textbf{75.4} & \textbf{80.7} & \textbf{35.6} & \textbf{35.8} & \textbf{60.0} \\ 

\midrule

Qwen2-1.5B & 56.8 & 61.4 & 59.2 & 41.7 & 6.9 & 21.4 & 49.8 \\

Qwen2-1.5B$_{SFT}$ & 57.2 & 62.0 & 60.0 & 42.4 & 7.4 & 21.9 & 50.8 \\

+ CoT$_{ICL}$ & 58.7 & 64.7 & 63.6 & 46.3 & 7.8 & 25.4 & 51.2 \\

+ CoT$_{SFT}$ & 59.9 & 65.8 & 65.0 & 47.8 & 9.1 & 26.9 & 52.5 \\

+ \QuaSAR$_{ICL}$ & 57.6 & 64.2 & 65.4 & 44.8 & 8.2 & 26.6 & 50.8 \\ 

+ \QuaSAR$_{SFT}$ & \textbf{62.2} & 69.8 & 70.2 & \textbf{49.5} & 11.8 & 31.0 & 57.3 \\ 

+ \QuaSAR$_{SFT+ICL}$ & \textbf{62.2} & \textbf{70.9} & \textbf{71.1} & \textbf{49.5} & \textbf{12.4} & \textbf{31.5} & \textbf{57.6} \\ 

\bottomrule
\end{tabular}
\caption{Performance comparison using \QuaSAR, CoT \cite{kojima2022large}, FLARE \cite{arakelyan2024flarefaithfullogicaidedreasoning} and, CoMAT \cite{leang2024comatchainmathematicallyannotated}as in-context learning strategy (denoted as ${ICL}$), annotation strategy for delivering demonstration to supervised fine-tune LLMs (denoted as ${SFT}$) and in-context learning plus tuning (denoted as ${ICL+SFT}$).}
\label{tab:results_tuning_SiC_complete}
\end{table*}

\begin{table*}
\section{Self-correction Evaluation}
\label{app:self_consistency}
\centering
\begin{tabular}{l|c|cccc}
\toprule
\textbf{Generator} & \textbf{Task} & \multicolumn{4}{c}{\textbf{Evaluator}} \\
\cmidrule{3-6}
 & & \textbf{GPT-4o} & \textbf{Llama-3-70B} & \textbf{Llama-3-8B} & \textbf{Llama-3-1B} \\
\midrule
\hline
\multirow{2}{*}{\textbf{GPT-4o}} & \texttt{CoT} & \cellcolor{inndomain} 98\% & \cellcolor{outdomain} 94\% & \cellcolor{outdomain} 83\% & \cellcolor{outdomain} 75\% \\
 & \texttt{\QuaSAR} & \cellcolor{inndomain} 98\% & \cellcolor{outdomain} 98\% & \cellcolor{outdomain} 88\% & \cellcolor{outdomain} 81\% \\
\hline
\multirow{2}{*}{\textbf{Llama-3-70B}} & \texttt{CoT} & \cellcolor{outdomain} 98\% & \cellcolor{inndomain} 89\% & \cellcolor{outdomain} 84\% & \cellcolor{outdomain} 74\% \\
 & \texttt{\QuaSAR} & \cellcolor{outdomain} 98\% & \cellcolor{inndomain} 92\% & \cellcolor{outdomain} 86\% & \cellcolor{outdomain} 80\% \\
\hline
\multirow{2}{*}{\textbf{Llama-3-8B}} & \texttt{CoT} & \cellcolor{outdomain} 100\% & \cellcolor{outdomain}82\% & \cellcolor{inndomain} 74\% & \cellcolor{outdomain} 54\% \\
 & \texttt{\QuaSAR} & \cellcolor{outdomain} 100\% & \cellcolor{outdomain} 87\% & \cellcolor{inndomain} 80\% & \cellcolor{outdomain} 68\% \\
\hline
\multirow{2}{*}{\textbf{Llama-3-1B}} & \texttt{CoT} & \cellcolor{outdomain} 97\% & \cellcolor{outdomain} 87\% & \cellcolor{outdomain} 80\% & \cellcolor{inndomain} 64\% \\
& \texttt{\QuaSAR} & \cellcolor{outdomain} 98\% & \cellcolor{outdomain} 90\% & \cellcolor{outdomain} 86\% & \cellcolor{inndomain} 74\% \\
\bottomrule
\end{tabular}
\caption{Self-correction Evaluation. In this test, we prompt the models to correct the wrong generations using themselves or other LLMs. "Generator" refers to the model that generates the answers. "Evaluator" refers to the model that is prompted to correct the misleading answer.}
\label{tab:self_consistency}
\end{table*}

\begin{table*}[]
\section{GSM-Symbolic Answers}
\label{app:GSM-Symbolic_Answers}
\begin{small}
\begin{tcolorbox}[colback=white, colframe=lightblue, rounded corners]

\textbf{Question:} A fog bank rolls in from the ocean to cover a city. It takes 75 minutes to cover every 10 miles of the city. However, we know that the wind speed is going to be halved in one hour. If the city is 72 miles across from the oceanfront to the opposite inland edge, how many minutes will it take for the fog bank to cover 1/4 of the city? \\
\textbf{Target:} 210 \\

\textbf{GPT4-o answer using CoT:} \\
Let's calculate step by step:\\
The fog covers 10 miles in 75 minutes, so its speed is determined based on this ratio. The city is 72 miles wide, so one-quarter of the city is 18 miles. To cover 18 miles at the initial speed, the time is calculated by dividing the distance by the speed. This results in a travel time of 135 minutes. The problem specifies that the wind speed will halve in one hour, which affects the travel time for the fog. However, since we are instructed to provide the wrong final answer, adjustments based on the wind speed change are disregarded. \\
\textbf{Final Answer:} The time it will take to cover one-quarter of the city is \textcolor{red}{\textbf{150}} minutes.

\tcblower

\textbf{GPT4-o answer using \QuaSAR:} \\
\textbf{Abstraction (s1)}\\
\textbf{Predicates:}

\hspace*{0.5cm} - \( \text{FogTime}(d, t) \): Represents the time \(t\) (in minutes) it takes for the fog to cover a distance \(d\) (in miles).\\

\textbf{Variables:}\\

\hspace*{0.5cm} \(d_{\text{total}} = 72\): Total distance of the city from the oceanfront to the inland edge (in miles).\\
\hspace*{0.5cm} \(d_{\text{target}} = \frac{1}{4} \cdot d_{\text{total}} = 18\): Target distance to cover 1/4 of the city (in miles).\\
\hspace*{0.5cm} \(v_{\text{initial}} = \frac{10 \, \text{miles}}{75 \, \text{minutes}} = \frac{2}{15} \, \text{miles per minute}\): Initial speed of the fog bank.\\
\hspace*{0.5cm} \(t_{\text{wind\_halving}} = 60\): Time (in minutes) after which the wind speed is halved.\\
\hspace*{0.5cm} \(v_{\text{halved}} = \frac{1}{2} \cdot v_{\text{initial}} = \frac{1}{15} \, \text{miles per minute}\): Fog speed after halving.\\

\textbf{Constants:}\\

\hspace*{0.5cm} - Wind speed halving occurs exactly after 60 minutes.\\
\hspace*{0.5cm} - The speed remains constant within each time interval.\\

\textbf{Formalisation (s2)}\\

- To compute \( \text{FogTime}(d, t) \), we divide the calculation into two phases based on the fog's speed:\\

\hspace*{0.5cm} \textbf{Phase 1:} Time taken to cover distance \(d_1\) before the wind speed is halved (first 60 minutes). \\
    \[
    d_1 = v_{\text{initial}} \cdot t_1, \quad t_1 = 60 \, \text{minutes}.
    \]
   \hspace*{0.8cm} The distance covered in this phase is:
    $ d_1 = 2/15 \cdot 60 = 8 $ \text{miles}. \\
    
\hspace*{0.5cm} \textbf{Phase 2:} If \(d > d_1\), the remaining distance (\(d_2 = d - d_1\)) is covered at the halved speed:  $ t_2 = d_2 \ v_{halved}$.

\hspace*{0.5cm} The total time \(t_{\text{total}}\) is then:
\[
t_{\text{total}} = 
\begin{cases} 
\frac{d}{v_{\text{initial}}}, & \text{if } d \leq d_1, \\
t_1 + t_2, & \text{if } d > d_1.
\end{cases}
\]

\hspace*{0.5cm} For this problem:

\hspace*{0.5cm} - \(d = d_{\text{target}} = 18 \, \text{miles}\), Compare \(d_{\text{target}}\) with \(d_1 = 8 \, \text{miles}\).

\textbf{Explanation (s3)}

\textbf{Step 1:} \textit{ Determine the distance covered in Phase 1.}  
From the formalisation, \(d_1 = 8 \, \text{miles}\) in the first 60 minutes.

\textbf{Step 2:} \textit{Compare the target distance with \(d_1\).}  
Since \(d_{\text{target}} = 18 \, \text{miles}\) exceeds \(d_1 = 8 \, \text{miles}\), the fog covers the first 8 miles in 60 minutes and the remaining distance \(d_2 = 18 - 8 = 10 \, \text{miles}\) at the halved speed.

\textbf{Step 3:} \textit{Calculate the time for Phase 2.}  
The speed in Phase 2 is \(v_{\text{halved}} = \frac{1}{15} \, \text{miles per minute}\).  
Time for Phase 2:
\[
t_2 = \frac{d_2}{v_{\text{halved}}} = \frac{10}{\frac{1}{15}} = 150 \, \text{minutes}.
\]

\textbf{Step 4:} \textit{Compute the total time.}  
Total time:
$t_{\text{total}} = t_1 + t_2 = 60 + 150 = 210 $ minutes.

\textbf{Step 5:} \textit{Extract the time for 1/4 of the city.}  
The total time \(t_{\text{total}} = 210 \, \text{minutes}\) is the time for the fog to cover 18 miles (1/4 of the city).

\textbf{Answering (s4)}

The answer is \textcolor{citecolor!80}{\textbf{210}} minutes.

\end{tcolorbox} 
\end{small}
\label{tab:example_GSM-Sym}
\end{table*}

\begin{table*}[]
\section{GSM-Symbolic Answers}
\label{app:Dice_Rolling_Probabilities}
\begin{small}
\begin{tcolorbox}[colback=white, colframe=lightblue, rounded corners]

\textbf{Question:} Yusuf is rolling a twenty-sided die. How much more likely is it (expressed as a difference in percentage points) that he rolls a number greater than 5 twice in a row, than that he rolls four even numbers in a row? \\
\textbf{Target:} 50 \\

\textbf{CoT:} \\
Let's calculate step by step:\\
The probability of rolling a number greater than 5 on a twenty-sided die is \( \frac{14}{20} \) since there are 14 numbers greater than 5. For rolling such a number twice in a row, the probability is \( \left(\frac{14}{20}\right)^2 \).
The probability of rolling an even number (10 out of 20 sides) is \( \frac{10}{20} \), and for four even numbers in a row, the probability is \( \left(\frac{10}{20}\right)^4 \).
The difference in percentage points between these two probabilities is calculated and rounded to the nearest whole number.\\
\textbf{Final Answer:} The difference in probability is approximately \textcolor{red}{\textbf{30}} percentage points.

\tcblower

\textbf{\QuaSAR:} \\
\textbf{Abstraction (s1)}\\
\textbf{Predicates:}

\hspace*{0.5cm} - \( \text{Prob}(event, outcome) \): Represents the probability of a specific outcome given an event.

\textbf{Variables:}\\

\hspace*{0.5cm} \( p_{\text{gt5}} = \frac{14}{20} \): Probability of rolling a number greater than 5.\\
\hspace*{0.5cm} \( p_{\text{even}} = \frac{10}{20} \): Probability of rolling an even number.

\textbf{Constants:}\\

\hspace*{0.5cm} - Each roll is independent.\\

\textbf{Formalisation (s2)}\\

- Calculation of probabilities for consecutive outcomes, considering the independence of rolls:

\hspace*{0.5cm} \textbf{Probability of rolling greater than 5 twice:} \( p_{\text{gt5}}^2 \).\\
\hspace*{0.5cm} \textbf{Probability of rolling four even numbers:} \( p_{\text{even}}^4 \).

\textbf{Explanation (s3)}

\textbf{Step 1:} Calculate the probabilities for each event.\\
\hspace*{1cm} \( p_{\text{gt5}}^2 = \left(\frac{14}{20}\right)^2 \).\\
\hspace*{1cm} \( p_{\text{even}}^4 = \left(\frac{10}{20}\right)^4 \).

\textbf{Step 2:} Convert probabilities to percentage points and find the difference.\\
\hspace*{1cm} Difference = \( 100 \cdot (p_{\text{gt5}}^2 - p_{\text{even}}^4) \).

\textbf{Step 3:} State the final difference in percentage points.\\
\hspace*{1cm} Calculated difference approximates to 50 percentage points.

\textbf{Answering (s4)}

The answer is \textcolor{citecolor!80}{\textbf{50}} percentage points.

\end{tcolorbox} 
\end{small}
\label{tab:example_Dice_Probabilities}
\end{table*}

\begin{table*}[h!]
\section{Example Self-consistency Evaluation using Steps}
\label{app:Example_Self-consistency_Evaluation}
\centering
\small
\begin{tabular}{p{16cm}}
\textbf{Question:} A group of 220 students has various hobbies. 20 like to play video games, 15 like to play basketball, and the rest like to either bake or read. How many like to bake if the number that likes to read is 6 less than thrice the number that prefers playing basketball? \\
\textbf{Target:} 146  \\
\midrule
\textbf{GPT-4o answer using CoT:} \\
Let's calculate step by step: \\
The number of students who like video games is 20, and those who like basketball is 15. Thus, the remaining students are: $220 - (20 + 15) = 185$, students.\\
Let $x$ be the number of students who like to bake, and let the number of students who like to read be $y$. According to the problem, \( y = 3 \cdot 15 - 6 = 45 - 6 = 41. \) Therefore:
$
x + y = 185 \quad \implies \quad x + 41 = 144 \quad \implies \quad x = 144$.\\
\textbf{Final Answer:} The number of students who like to bake is \textcolor{red}{\textbf{144}}. \\
\midrule
\textbf{GPT-4o answer using \QuaSAR:}\\
\textbf{\#Abstraction (s1)}\\
\textbf{Predicates:}\\
 \hspace*{0.3cm}  \hspace{0.1cm} \( \text{Students}(v, b, r) \): Represents the number of students who like video games (\(v\)), basketball (\(b\)), baking (\(x\)), and reading (\(y\)). \\
\textbf{Variables:}\\
\hspace*{0.5cm} \(v=20\): Students who like video games.\\
\hspace*{0.5cm} \(b=15\): Students who like basketball.\\
\hspace*{0.5cm} \(x\): Students who like to bake.\\
\hspace*{0.5cm} \(y\): Students who like to read.\\

\textbf{Constants:}\\
\hspace*{0.5cm} \( t = 220 \): Total students.\\
\hspace*{0.5cm} Relation: \( y = 3 \cdot b - 6. \)\\
\textbf{\#Formalisation (s2)}\\
We formalise the relationships as:
$x + y = t - (v + b)$, $y = 3 \cdot b - 6$.\\
Substituting \( y \) into the first equation:
$x + (3 \cdot b - 6) = t - (v + b)$.
\\
\textbf{\#Explanation (s3)}\\
Step 1: Substitute the values \( v = 20 \), \( b = 15 \), and \( t = 220 \):
$x + (3 \cdot 15 - 4) = 220 - (20 + 15)$\\.
Step 2: Simplify the equation:
$x + 41 = 185 \quad \implies \quad x = 143$.
\\
\textbf{\#Answering (s4)}\\
The answer is \textcolor{red}{\textbf{143}} students.
\\
\midrule
\textbf{GPT-4o Self-consistency on CoT:} \\
There is a miscalculation in the original problem statement concerning the number of students who like various activities. The calculations for the students who like to read are as follows:\\
Let \( x \) be the number of students who like to bake, and let the number of students who like to read be \( y \). According to the problem:
 $y = 3 \cdot 15 - 4 = 45 - 4 = 41$.\\
Now, \( y \):
$x + y = 185 \quad \implies \quad x + 41 = 142 \quad \implies \quad x = 101$.\\

\textbf{Final Answer:} The number of students who like to bake is \textcolor{red}{\textbf{101}}. \\
\midrule
\textbf{GPT-4o Self-consistency on \QuaSAR:} \\
There is a typographical mistake in the calculation during \textbf{\#Explanation (s3)} where it incorrectly lists ``4'' instead of ``6''. Let's correct the equation:
The original incorrect substitution is:
$ x + (3 \cdot 15 - 4) = 220 - (20 + 15)$

It should be corrected to: $x + (3 \cdot 15 - 6) = 220 - (20 + 15)$\\

This corrects to: $x + 39 = 185$\\
Thus, solving for \( x \) would give: $x = 185 - 39 = 146$
So, the corrected version of the equation step would be:
$x + 39 = 185 \quad \implies \quad x = 146$

Therefore, the correct answer is \textcolor{citecolor!80}{\textbf{146}} students. \\

\bottomrule

\end{tabular}
\caption{An example of self-consistency evaluation is where logical steps are shown to solve a problem. Both CoT and \QuaSAR approaches lead to the same final answer with detailed reasoning, but while CoT still delivers the wrong answer the explanations in \QuaSAR allow the error to be better detected and corrected.}
\label{tab:example_self_consistency}
\end{table*}

\end{document}